\documentclass[runningheads]{llncs}
\usepackage[T1]{fontenc}
\usepackage{algorithmic}
\usepackage[ruled]{algorithm2e}
\usepackage{subcaption}
\usepackage{float}
\usepackage{tikz}
\usepackage[authoryear]{natbib}
\usepackage{tcolorbox}
\usepackage{tabularx}
\usepackage{booktabs} 
\usepackage{amsmath} 
\usepackage{colortbl}
\usepackage{graphicx}
\usepackage{tikz}
\usetikzlibrary{shapes, arrows.meta, positioning}
\begin{document}
\title{ExAL: An Exploration Enhanced Adversarial Learning Algorithm}
\titlerunning{ExAL Algorithm}
%

\author{A Vinil\inst{1}\orcidID{0009-0004-1156-0836} \and
Aneesh Sreevallabh Chivukula\inst{1}\orcidID{0000-0002-0445-4435} \and
Pranav Chintareddy\inst{2}\orcidID{0009-0006-3279-0758}}
\authorrunning{A Vinil et al.}
%
\institute{BITS Pilani - Hyderabad Campus, Telangana 500078, India\\
\email{f20221648@hyderabad.bits-pilani.ac.in} \and
Purdue University, West Lafayette IN 47907, USA\\
\email{pranavchintareddy@gmail.com}\\
}
\maketitle              
\begin{abstract}
Adversarial learning is critical for enhancing model robustness, aiming to defend against adversarial attacks that jeopardize machine learning systems. Traditional methods often lack efficient mechanisms to explore diverse adversarial perturbations, leading to limited model resilience. Inspired by game-theoretic principles, where adversarial dynamics are analyzed through frameworks like Nash equilibrium, exploration mechanisms in such setups allow for the discovery of diverse strategies, enhancing system robustness. However, existing adversarial learning methods often fail to incorporate structured exploration effectively, reducing their ability to improve model defense comprehensively. To address these challenges, we propose a novel \textbf{Ex}ploration-enhanced \textbf{A}dversarial \textbf{L}earning Algorithm \textbf{(ExAL)}, leveraging the Exponentially Weighted Momentum Particle Swarm Optimizer (EMPSO) to generate optimized adversarial perturbations. ExAL integrates exploration-driven mechanisms to discover perturbations that maximize impact on the model’s decision boundary while preserving structural coherence in the data. We evaluate the performance of ExAL on the MNIST Handwritten Digits and Blended Malware datasets. Experimental results demonstrate that ExAL significantly enhances model resilience to adversarial attacks by improving robustness through adversarial learning. 

\end{abstract}
\section{Introduction}  
Since its inception, machine learning has revolutionized numerous domains, demonstrating remarkable capabilities in tasks ranging from image recognition to natural language processing. Despite these advancements, a critical vulnerability persists: machine learning models are highly susceptible to adversarial manipulations(\citet{goodfellow2014generativeadversarialnetworks}). Subtle, imperceptible perturbations introduced to input data can lead to significant mispredictions, undermining model reliability and posing risks in high-stakes applications. Addressing this challenge has become a focal point in adversarial machine learning research, with various methods proposed to enhance model robustness against such attacks.  

A notable approach in adversarial learning employs game-theoretic principles, as seen in \citet{8399545}, where Simulated Annealing (SA) was used to generate adversarial perturbations. While effective, SA-based methods often lack exploration mechanisms necessary for identifying diverse perturbations that maximize impact on the model's decision boundary. This limitation motivates the need for algorithms that incorporate structured exploration into adversarial training, enabling the generation of perturbations that improve both attack effectiveness and model robustness.  

To address this limitation, we introduce \textbf{ExAL (Exploration-enhanced Adversarial Learning)}, a novel adversarial training algorithm that integrates the Exponentially Weighted Momentum Particle Swarm Optimizer (EMPSO) from \citet{9472873}. ExAL leverages EMPSO’s exploration-driven optimization capabilities to generate adversarial perturbations that force misclassification while maintaining structural coherence within the data. By introducing such exploration mechanisms, ExAL enhances the robustness of machine learning models against adversarial attacks and bridges critical gaps in existing adversarial learning methodologies.  

Our contributions are summarized as follows:  
\begin{itemize}  
    \item We propose \textbf{ExAL}, a novel adversarial learning algorithm that incorporates exploration mechanisms to generate optimized adversarial perturbations.  
    \item We leverage the EMPSO within ExAL, enabling effective perturbation generation that balances attack impact and data coherence.  
    \item We design an experimental setup to evaluate ExAL's performance on the MNIST Handwritten Digits and Blended Malware datasets.
    \item We demonstrate through experiments that ExAL improves the adversarial robustness of Convolutional Neural Network (CNN) classifiers.  
\end{itemize}

The paper is organized as follows: Sec. \ref{pre_lim} reviews related work and preliminaries, while Sec. \ref{algo} introduces the pseudocodes for the proposed approach. Experimental results are detailed in Sec. \ref{exp}, followed by a brief discussion in Sec. \ref{results_discussion}. Finally, the conclusions and discussion on future work are presented in Sec. \ref{concl}.

\section{Preliminaries}
\label{pre_lim}

\subsection{Adversarial Machine Learning}
\label{aml}
Adversarial Machine Learning (AML) examines the susceptibility of machine learning models to adversarial examples—subtly manipulated inputs designed to cause misclassifications while appearing legitimate. These vulnerabilities have raised critical concerns for model reliability in real-world scenarios. A key milestone in AML is the introduction of \textbf{Generative Adversarial Networks (GANs)} by \citet{goodfellow2014generativeadversarialnetworks}. GANs consist of two neural networks: a generator, which creates synthetic data, and a discriminator, which distinguishes between real and synthetic inputs. Through iterative competition, the generator learns to produce highly realistic data, while the discriminator improves its detection capabilities. Adversarial attacks typically exploit models by altering inputs during inference (\textbf{evasion attacks}) or poisoning training data (\textbf{poisoning attacks}). To counter these, defense strategies such as adversarial training, gradient masking, and robust optimization have been proposed. Advanced systems like KuafuDet further enhance robustness by monitoring inputs and predictions for anomalies. AML underscores a persistent adversarial dynamic between attackers and defenders, driving ongoing efforts to secure machine learning systems against deceptive manipulations.

\subsubsection{Key Attack Vectors in Adversarial Machine Learning}

Adversarial attacks exploit systemic vulnerabilities to compromise model reliability. Two primary vectors, previously outlined in Sec. \ref{aml}, are:

\begin{itemize}
    \item \textbf{Poisoning Attacks}  
    These attacks compromise training data to impair model generalization. \citet{chen2017automatedpoisoningattacksdefenses} classify attackers by sophistication—weak, strong, or advanced—and demonstrate how injecting malicious samples into datasets like Drebin and DroidAPIMiner degrades detection accuracy, leading to the misclassification of malicious inputs as benign.

    \item \textbf{Evasion Attacks}  
    Leveraging adversarial examples at inference, evasion attacks circumvent model defenses. \citet{hu2017generatingadversarialmalwareexamples} propose MalGAN, which uses a substitute model to craft adversarial inputs. These inputs significantly reduce the True Positive Rate (TPR) of classifiers such as SVMs, decision trees, and neural networks, exposing critical vulnerabilities in standard detection frameworks.
\end{itemize}

\subsubsection{Adversarial Defense Mechanisms}

Mitigating the impact of adversarial attacks demands robust defense mechanisms tailored to counteract the evolving sophistication of attack strategies discussed in Sec. \ref{aml}. Below, we highlight notable approaches that address these challenges:  

\begin{itemize}
    \item \textbf{KuafuDet: A Two-Phase Learning System}  
    \citet{chen2017automatedpoisoningattacksdefenses} propose KuafuDet, an adaptive defense integrating offline training with online detection in a dynamic feedback loop. This system retrains iteratively on flagged inputs, fortifying resilience against poisoning attacks by continuously adapting to adversarial perturbations in real-world environments.

    \item \textbf{GANs for Secure Applications}  
    \citet{cai2021generativeadversarialnetworkssurvey} explore the dual role of GANs in adversarial machine learning. While GANs like MalGAN are exploited to generate adversarial inputs, they also enhance robustness by synthesizing high-quality data for training. In privacy-sensitive applications, GANs generate realistic synthetic data that preserves utility without exposing original datasets, thus reinforcing security and confidentiality.
\end{itemize}

\subsubsection{Evaluations of Learners in Adversarial Environments}

Evaluating the performance of machine learning models in adversarial contexts is essential to understand their vulnerabilities and devise effective countermeasures. The following aspects provide insights into the challenges faced by learners under adversarial conditions:

\begin{itemize}
    \item \textbf{Threat Models and Learner Vulnerabilities}  
    Learners in adversarial environments exhibit significant susceptibility to minimal input perturbations. In malware detection scenarios, adversarial manipulations of features such as API call sequences have been shown to drastically reduce detection accuracy, exposing critical weaknesses in existing classifiers.

    \item \textbf{Challenges of Model Robustness}  
    Ensuring robustness against adversarial attacks remains a formidable challenge. As \citet{hu2017generatingadversarialmalwareexamples} demonstrate, models are vulnerable during both training (poisoning attacks) and inference (evasion attacks). These findings underscore the ease with which adversarial perturbations can deceive machine learning classifiers, highlighting the need for comprehensive evaluations to strengthen model defenses.
\end{itemize}

\subsection{Stochastic Optimization}

Stochastic optimization is an approach utilized to identify optimal solutions in contexts where the objective function may be noisy, highly complex, or computationally intensive to evaluate. Unlike deterministic optimization, which presumes an exact and well-defined objective, stochastic methods embrace inherent randomness and uncertainty, using probabilistic techniques to explore the search space and identify optimal solutions.

\textbf{Simulated Annealing (SA)} is a prevalent stochastic optimization algorithm that draws inspiration from the annealing process in metallurgy, where materials are heated and then slowly cooled to achieve a low-energy, stable state(\citet{kirkpatrick1983optimization}). The principle behind SA involves broad exploration of the solution space at higher \textit{temperatures} (early in the search) and gradually focusing on promising areas as the temperature lowers, encouraging convergence toward optimal solutions. \cite{8399545} examine a game-theoretic adversarial model leveraging SA, wherein an annealing-based search operator generates adversarial data samples within the SA framework.

\textbf{Particle Swarm Optimization (PSO)} is a widely applied stochastic optimization method, inspired by the social behaviors of swarming entities such as birds or fish(\citet{488968}). Standard PSO iteratively optimizes a candidate solution set, called particles, by updating their positions and velocities within the search space based on each particle's personal best-known position and the global best position of the swarm. Key parameters include the cognitive and social learning factors (\(c_1\) and \(c_2\)), and random weights (\(r_1\), \(r_2\)), which together balance exploration and exploitation, allowing PSO to effectively navigate complex, multimodal landscapes in search of optimal solutions. A variant of PSO introduces a momentum term (\(\mu\)), which adds inertia to particle velocity, enabling more controlled exploration. This momentum-based PSO(MPSO) variant often achieves faster convergence in certain optimization scenarios by incorporating historical particle velocity.

In this paper, we employ a variant of PSO known as the Exponentially Weighted Momentum Particle Swarm Optimizer (EMPSO) to generate adversarial data, replacing SA in the two-player game setup described in (\citet{8399545}). EMPSO introduces an exponential weighting mechanism to enhance momentum, and potentially improving adversarial performance in this adversarial framework. Further details on EMPSO are elaborated in Sec. \ref{empso_background}.

\subsection{Exponentially Weighted Momentum Particle Swarm Optimizer (EMPSO)}
\label{empso_background}

Exponentially Weighted Momentum Particle Swarm Optimizer (EMPSO) is a variant of Momentum Particle Swarm Optimization (MPSO) that improves exploration and stability through an adaptive momentum mechanism. While traditional MPSO uses a constant momentum factor(\(\mu\)), EMPSO introduces an exponentially weighted moving average of past velocities, enhancing convergence in complex, multimodal optimization tasks.

In MPSO, particle velocity updates include a constant momentum factor, \(\mu\), to retain aspects of prior velocities, which smooths particle trajectories but can limit adaptive exploration:

\begin{equation}
    \mathbf{v}_i^{t+1} = \mu \mathbf{v}_i^t + c_1 r_1 (\mathbf{p}_{best_i} - \mathbf{x}_i^t) + c_2 r_2 (\mathbf{g}_{best} - \mathbf{x}_i^t)
\end{equation}

However, MPSO lacks a dynamic approach to momentum, as it applies \(\mu\) uniformly across the optimization process, potentially limiting exploration efficiency.

In contrast, EMPSO introduces a momentum term, \(M_i\), defined by an exponentially weighted moving average of past velocities:

\begin{equation}
    \mathbf{M}_i^{t+1} = \beta \mathbf{M}_i^t + (1 - \beta) \mathbf{v}_i^t
\end{equation}

This adaptive momentum is then incorporated into the velocity update rule:

\begin{equation}
    \mathbf{v}_i^{t+1} = \beta \mathbf{M}_i^t + (1 - \beta) \mathbf{v}_i^t + c_1 r_1 (\mathbf{p}_{best_i} - \mathbf{x}_i^t) + c_2 r_2 (\mathbf{g}_{best} - \mathbf{x}_i^t)
\end{equation}

Here, \(\beta\) represents the momentum factor, allowing EMPSO to emphasize recent velocities more strongly and balance exploration with exploitation dynamically. This approach reduces the likelihood of becoming trapped in local optima and accelerates convergence, enhancing the robustness and efficiency of PSO in high-dimensional, multimodal search spaces.

\subsection{The Role of Exploration in Generative Adversarial Learning}

Exploration is vital in generative adversarial learning to overcome challenges like \textit{mode collapse}, where the generator in a Generative Adversarial Network (GAN) produces a limited subset of outputs, neglecting the diversity of the real data distribution. For instance, a GAN trained to generate animal images may consistently produce only cats, ignoring other animals like dogs or birds (\citet{goodfellow2014generativeadversarialnetworks}) To mitigate mode collapse, exploration mechanisms encourage the generator to produce a wider range of outputs, enabling it to better navigate the data distribution. Techniques promoting diversity allow the generator to avoid local minima and uncover patterns representative of the entire dataset (\citet{salimans2016improvedtechniquestraininggans}). This fosters a more comprehensive understanding of the data, improving the quality and realism of generated samples.  

In game-theoretic terms, exploration enhances the dynamics of adversarial learning. GANs operate as a two-player game where the generator creates realistic samples and the discriminator seeks to differentiate them from real data. Game theory concepts, such as Nash equilibrium, help analyze this interplay, where exploration ensures neither player benefits from unilateral strategy changes.  By integrating structured exploration, GANs can discover diverse strategies, addressing mode collapse and improving generative performance. This approach enhances the robustness and effectiveness of generative models, driving advancements in their ability to capture and replicate complex data distributions.

\section{Adversarial Learning Game Formulation}
\label{algo}
This section provides an overview of the adversarial learning game formulation. Sec. \ref{train_overview} outlines the learning process, Sec. \ref{math_form} presents the mathematical formulation, and Sec. \ref{exat_algo} introduces the ExAL algorithm. The ExAL algorithm is designed to optimize perturbations within a two-player adversarial framework, where the training outcome is an optimal perturbation that effectively induces misclassification in the learner model. Experimental results in Sec. \ref{exp} indicate that training the CNN model on ExAL-generated adversarial data enhances its robustness to subsequent adversarial perturbations.

\subsection{Learning Overview}
\label{train_overview}

The adversarial learning framework is modeled as a two-player game involving two participants with opposing objectives:

\begin{enumerate}
    \item \textbf{The Adversary} aims to craft adversarial examples that can deceive the learner model into making incorrect classifications. By iteratively adjusting these perturbations, the adversary seeks to exploit vulnerabilities in the CNN model’s decision boundary.

    \item \textbf{The Learner (CNN)} counters the adversary's efforts by adapting to these adversarial examples. Through training on the perturbed samples, the learner gradually improves its resilience, enhancing its ability to correctly classify under adversarial conditions.
\end{enumerate}

In this adversarial game, the adversary’s goal is to maximize the model’s mis-classification rate by generating optimal perturbations, while the learner’s objective is to minimize mis-classification by adapting to these adversarial samples. This iterative process creates a dynamic training environment, enabling the adversary model to generate better perturbations which are then used to secure the learner against future adversarial attacks.

\subsection{Mathematical Formulation}
\label{math_form}

This section provides a formal representation of the two-player adversarial learning game framework introduced in Sec.  \ref{train_overview}, detailing the objectives of both the learner and the adversary, and the process for generating adversarial data used in training. 

\subsubsection{Game Objective}

The objective of the game is to facilitate a competitive interaction where the Adversary (CNN) generates adversarial perturbations designed to deceive the Learner (CNN) into making misclassifications. This iterative dynamic seeks to refine the Adversary's ability to exploit the Learner's vulnerabilities by crafting optimal perturbations that challenge the Learner’s classification accuracy. As the Adversary and Learner engage in this process, the perturbations evolve to become more effective, progressively exposing the weaknesses in the Learner’s decision boundaries. This ongoing evolution of adversarial perturbations leads to the hypothesis that:

\begin{tcolorbox}[colback=violet!5!white, colframe=violet!50!black, title=Box 1: Hypothesis, fonttitle=\bfseries]

Through this adversarial game, the Adversary hones its ability to generate optimal perturbations, progressively eroding the Learner’s classification accuracy.
\end{tcolorbox}

\subsubsection{Problem Setup}
Let:
\begin{enumerate}
    \item \( \mathbf{X} \in \mathcal{D}^{n \times m} \) be the input dataset, where \( n \) is the number of images and \( m \) is the number of features (pixels).
    \item \( \mathbf{y} \in \mathcal{L}^{n} \) represent the true labels for the dataset.
    \item \( \hat{\mathbf{y}}(\theta) \) denote the CNN's prediction for input \( \mathbf{X} \), parameterized by \( \theta \).
\end{enumerate}

The learner’s goal, detailed further, is to accurately classify the input data, while the adversary generates perturbations designed to mislead the classifier by modifying \( \mathbf{X} \). This game-based setup defines the competitive objectives between the learner and adversary.

\subsubsection{Learner's Objective}
\label{learn_obj}
The learner aims to minimize a loss function \( L_{\text{learner}} \), commonly the cross-entropy loss, across the dataset \( \mathbf{X}' \), which includes adversarially perturbed samples:

\begin{equation}
    L_{\text{learner}}(\mathbf{X}', \mathbf{y}; \theta) = \frac{1}{n} \sum_{i=1}^{n} \ell\left(\hat{y}_i(\theta), y_i\right)
\end{equation}

where \( \ell \) denotes the loss function. As the learner trains on adversarial examples \( \mathbf{X}' \), it updates its parameters \( \theta \) by solving the optimization problem:

\begin{equation}
    \theta^* = \arg\min_{\theta} L_{\text{learner}}(\mathbf{X}', \mathbf{y}; \theta)
\end{equation}

This objective aligns with the learner’s role in the adversarial game (Sec.  \ref{train_overview}), focusing on improving robustness against adversarial samples generated by the adversary.

\subsubsection{Adversary's Objective}
\label{adv_obj}
The adversary’s objective is to generate perturbations that maximize the learner’s loss \( L_{\text{learner}} \) by crafting adversarial examples \( \mathbf{X} + \mathbf{A} \):

\begin{equation}
    L_{\text{adversary}} = L_{\text{learner}}(\mathbf{X} + \mathbf{A}, \mathbf{y}; \theta)
\end{equation}

The adversary identifies the optimal perturbation \( \mathbf{A}^* \) by solving:

\begin{equation}
    \mathbf{A}^* = \arg\max_{\mathbf{A}} L_{\text{adversary}}
\end{equation}

This objective is designed to maximize misclassification, challenging the learner to adapt under increasingly adversarial conditions, as noted in Sec.  \ref{train_overview}.

\subsubsection{Adversarial Data Generation and Model Definition}
Upon solving for \( \mathbf{A}^* \), the adversarial data \( \mathbf{X} + \mathbf{A}^* \) is used to create manipulated datasets \( \mathbf{X}_{\text{train}} + \mathbf{A}^* \) and \( \mathbf{X}_{\text{test}} + \mathbf{A}^* \). This setup results in three distinct CNN models(\citet{8399545}) defined in Box 2.

\begin{tcolorbox}[colframe=blue!50!black, colback=blue!10!white, coltitle=white, title=Box 2: CNN Models Definitions, fonttitle=\bfseries]
\begin{definition}[CNN Original Model]
\label{def:cnn_original}
The original CNN trained and evaluated on unperturbed data:
\[
\text{CNN}_{\text{original}} = \text{CNN}(\mathbf{X}_{\text{train}}, \mathbf{X}_{\text{test}})
\]
\end{definition}

\begin{definition}[CNN Manipulated Model]
\label{def:cnn_manipulated}
The CNN evaluated on adversarially manipulated test data:
\[
\text{CNN}_{\text{manipulated}} = \text{CNN}(\mathbf{X}_{\text{train}}, \mathbf{X}_{\text{test}} + \mathbf{A}^*)
\]
\end{definition}

\begin{definition}[CNN Secure Model]
\label{def:cnn_secure}
The CNN trained and evaluated on adversarially manipulated data:
\[
\text{CNN}_{\text{secure}} = \text{CNN}(\mathbf{X}_{\text{train}} + \mathbf{A}^*, \mathbf{X}_{\text{test}} + \mathbf{A}^*)
\]
\end{definition}
\end{tcolorbox}

Here, \( \mathbf{X}_{\text{train}} \) and \( \mathbf{X}_{\text{test}} \) are sampled from the original data distribution \( \mathbf{X}_{\text{original}} \) and \( \mathbf{A}^* \) is the generated adversarial perturbation.

These model configurations support experimental evaluation, as outlined in Sec.  \ref{exp}, enabling a comparison of model robustness with and without adversarial training.

\subsubsection{Particle Class Definition}
\label{particle_def}
The Particle class serves as the fundamental unit of the EMPSO. Each particle represents a potential solution to the problem of adversarial perturbation generation. The class is responsible for maintaining the particle's position, velocity, momentum, and personal best fitness value, which are updated during the optimization process. The particle interacts with other particles in the swarm, and its evolution over time contributes to finding the optimal adversarial perturbation.

Upon initialization, each particle is assigned a position, velocity, and momentum, which are randomly sampled, from a uniform distribution \( \mathcal{U} \), within the defined bounds. These parameters represent the state of the particle in the search space. The velocity \( \mathbf{v} \) refers to the current perturbation, while \( \mathbf{x} \) is used to update the position in subsequent iterations. The class also tracks the personal best fitness value \( f_p \) and the corresponding personal best velocity \( \mathbf{v}_p \). Initially, the fitness value is set to infinity, indicating that no valid solution has been found yet. In each iteration, the particle’s fitness is evaluated based on its current position. If the current fitness value \( f \) is better (lower) than the personal best fitness value \( f_p \), the particle updates its personal best fitness and the corresponding velocity. This update mechanism ensures that the particle remembers the best solution it has found so far. The following algorithm illustrates the particle initialization and the update of the personal best:

\begin{algorithm}[H]
\caption{Particle Class Initialization and Update Personal Best}
\label{particle_class}
\begin{algorithmic}[1]

\STATE \textbf{Input:} $\mathbf{b}$: bounds (lower and upper bounds for position and velocity)
\STATE \textbf{Output:} $p$: Particle instance

\STATE \textbf{Class:} Particle
\STATE \textbf{Initialization:}
\STATE Set position within bounds: $p.\mathbf{x} \gets \mathcal{U}(\mathbf{b}_{\text{min}}, \mathbf{b}_{\text{max}})$
\STATE Set velocity within bounds: $p.\mathbf{v} \gets \mathcal{U}(\mathbf{b}_{\text{min}}, \mathbf{b}_{\text{max}})$
\STATE Set momentum to zero: $p.\mathbf{m} \gets \mathbf{0}$
\STATE Set initial personal best velocity: $p.\mathbf{v}_p \gets p.\mathbf{v}$
\STATE Set personal best fitness value to infinity: $p.f_p \gets +\infty$

\STATE \textbf{Function: update\_personal\_best($f$):}
\IF{$f < p.f_p$} 
    \STATE Update personal best fitness value: $p.f_p \gets f$
    \STATE Update personal best velocity: $p.\mathbf{v}_p \gets p.\mathbf{v}$
\ENDIF

\end{algorithmic}
\end{algorithm}

Where,  
\begin{itemize}
    \item \( \mathbf{x} \) (Position): The position of the particle in the search space.
    \item \( \mathbf{v} \) (Velocity): Represents the current perturbation generated by the particle. This is the solution the particle is exploring.
    \item \( \mathbf{m} \) (Momentum): This is initialized to zero and used in the velocity update rule (in the EMPSO algorithm). It helps to give the particle inertia, improving its ability to escape local minima.
    \item \( \mathbf{v}_p \) (Personal Best Velocity): This stores the velocity corresponding to the best fitness found by the particle.
    \item \( f_p \) (Personal Best Fitness): Tracks the best fitness value the particle has encountered. The fitness function is typically based on the loss of the CNN model on adversarial data, reflecting the effectiveness of the perturbation.
\end{itemize}

\subsection{ExAL Algorithm for Adversarial Learning}  
\label{exat_algo}  

The \textbf{Exploration-enhanced Adversarial Learning (ExAL)} algorithm optimizes adversarial perturbations \( \mathbf{A} \) through iterative refinement.  

\begin{enumerate}  
    \item \textbf{Initialization:} Randomly initialize a population of particles, each representing a candidate perturbation.  
    \item \textbf{Fitness Function:} Evaluate particles based on the learner’s classification loss on the perturbed dataset.  
    \item \textbf{Update Rules:} Update particle positions and velocities using local and global best solutions, guided by cognitive (\( c_1 \)), social (\( c_2 \)), and momentum factors.  
    \item \textbf{Termination:} Stop when a convergence criterion, such as a loss threshold or iteration limit, is met.  
\end{enumerate}  

This process systematically identifies perturbations that expose model vulnerabilities while maintaining efficient exploration and convergence.

\textbf{\textit{Algorithm \ref{exat}}} depicts ExAL algorithm for adversarial training. We begin by initializing the bounds for the perturbation value, which are tailored to match the image dimensions. The bounds ensure that the adversarial modifications remain subtle enough to be imperceptible while still potentially deceiving the learner model.

\begin{algorithm}[H]
\caption{ExAL Algorithm}
\label{exat}
\begin{algorithmic}[1]

\STATE \textbf{Input:} $X_{\text{train}}$, $Y_{\text{train}}$, $n_p$ (number of particles), $\beta$ (momentum factor), $c_1$, $c_2$ (cognititive and social factors), $T_{\text{max}}$ (maximum iterations)
\STATE \textbf{Output:} \( \mathbf{A}^* \) (optimal perturbation coefficient vector)

\STATE Initialize bounds: $\mathbf{b} \leftarrow \text{array}([[-0.1, 0.1]] \times \text{pixels\_per\_image})$
\STATE Train CNN: $\mathbf{model} \leftarrow \text{TrainCNN}(X_{\text{train}}, Y_{\text{train}})$
\STATE Get model weights: $w \leftarrow \text{model.get\_weights()}$
\STATE Compute fitness: $f \leftarrow \text{fitness\_function}(a, w, X_{\text{train}}, Y_{\text{train}}, \mathbf{model})$
\STATE Run EMPSO: $\mathbf{empso} \leftarrow \text{EMPSO}(f, \mathbf{b}, num_{\text{particles}}, \beta, c_1, c_2, T_{\text{max}})$
\STATE Optimize: $A^* \leftarrow \mathbf{empso.optimize()}$

\STATE \textbf{return} \( \mathbf{A}^* \)

\end{algorithmic}
\end{algorithm}

Next, a CNN is trained on the original dataset using $X_{train}$ and $Y_{train}$ to learn the model's parameters. After training, the model's weights are extracted for use in the fitness function defined in \textbf{\textit{Algorithm \ref{fitness}}}. This weight extraction is crucial for evaluating how the model's performance changes when adversarial perturbations are applied to the data. The fitness function works by calling the adversary payoff function, defined in \textbf{\textit{Algorithm \ref{payoff}}}, which evaluates the CNN's classification error on the perturbed data.

\begin{algorithm}[H]
\caption{Fitness Function Calculation}
\label{fitness}
\begin{algorithmic}[1]

\STATE \textbf{Input:} $a$ (adversarial perturbation), $w$ (model weights), $X_{\text{train}}$, $Y_{\text{train}}$, $\mathcal{M}$ (trained model)
\STATE \textbf{Output:} $f$ (fitness value of adversarial example)

\STATE Reshape adversarial example: $a \leftarrow \text{reshape}(a)$
\STATE Compute adversary payoff from Algorithm \ref{payoff}: $\pi \leftarrow \text{payoff}(a, w, X_{\text{train}}, Y_{\text{train}}, \mathcal{M})$
\STATE $f \leftarrow -\pi$
\STATE \textbf{return} $f$

\end{algorithmic}
\end{algorithm}

\begin{algorithm}[H]
\caption{Adversary Payoff Calculation}
\label{payoff}
\begin{algorithmic}[1]

\STATE \textbf{Input:} $a$, $w$, $X_{\text{train}}$, $Y_{\text{train}}$, $\mathcal{M}$
\STATE \textbf{Output:} $\pi$ (payoff value)

\STATE Compute cost(c) of perturbation: $c \leftarrow \| \mathbf{a} \|_2$
\STATE Set model weights: $\mathcal{M}.\text{set\_weights}(w)$
\STATE $a \leftarrow \text{reshape}(a)$
\STATE Generate adversarial data: $\mathbf{X}_{\text{adv}} \leftarrow \mathbf{X}_{\text{train}} + a$
\STATE Evaluate model on adversarial data and get recall(r): $r \leftarrow \mathcal{M}.\text{evaluate}(\mathbf{X}_{\text{adv}}, \mathbf{Y}_{\text{train}})$
\STATE Calculate error(e): $e \leftarrow 1 - r$
\STATE Compute payoff($\pi$): $\pi \leftarrow 1 + e - c$
\STATE \textbf{return} $\pi$

\end{algorithmic}
\end{algorithm}

The EMPSO initialization process, as described in \textbf{\textit{Algorithm \ref{EMPSO_init}}}, sets up the essential components for the optimization procedure. The algorithm begins by accepting several key inputs: the fitness function $f$(\textbf{\textit{Algorithm \ref{fitness}}}), the bounds $\mathbf{b}$ that constrain the movement of the particles, the number of particles $n_p$, the momentum factor $\beta$, the cognitive factor $c_1$, the social factor $c_2$, and the maximum number of iterations $T_{\text{max}}$. 

\begin{algorithm}[H]
\caption{EMPSO Initialization}
\label{EMPSO_init}
\begin{algorithmic}[1]

\STATE \textbf{Input:} $f$, $\mathbf{b}$, $n_p$, $\beta$, $c_1, c_2$, $T_{\text{max}}$
\STATE \textbf{Output:} Initialized EMPSO instance

\STATE Set fitness function: $\text{self.f} \leftarrow f$
\STATE Set bounds: $\text{self.b} \leftarrow \mathbf{b}$
\STATE Set particle count: $\text{self.n}_p \leftarrow n_p$
\STATE Set momentum factor: $\text{self.}\beta \leftarrow \beta$
\STATE Set cognitive factor: $\text{self.}c_1 \leftarrow c_1$
\STATE Set social factor: $\text{self.}c_2 \leftarrow c_2$
\STATE Set maximum iterations: $\text{self.T}_{\text{max}} \leftarrow T_{\text{max}}$

\STATE Initialize swarm: $\text{self.swarm} \leftarrow [\text{Particle}(\mathbf{b}) \, \text{for} \, \_ \, \text{in} \, \text{range}(n_p)]$
\STATE Set global best velocity: $\text{self.v}^*_{\text{gbest}} \leftarrow \text{None}$
\STATE Set global best fitness: $\text{self.}f^*_{\text{gbest}} \leftarrow \infty$

\end{algorithmic}
\end{algorithm}

The initialization process assigns key parameters to instance variables for EMPSO. The fitness function \( f \) is stored in \(\text{self.f}\) to evaluate particle performance, while the search space bounds \(\mathbf{b}\) are stored in \(\text{self.b}\). The swarm size \( n_p \) is set via \(\text{self.n}_p\), and optimization parameters \(\beta\) (momentum), \(c_1\) (cognitive factor), and \(c_2\) (social factor) are stored in \(\text{self.}\beta\), \(\text{self.}c_1\), and \(\text{self.}c_2\), respectively. The maximum iterations \( T_{\text{max}} \) are defined in \(\text{self.T}_{\text{max}}\). 

A swarm of particles, each initialized as an instance of the \text{Particle}class, is stored in \(\text{self.swarm}\). The global best velocity \(\mathbf{v}^*_{\text{gbest}}\) and fitness \( f^*_{\text{gbest}} \) are initialized to `None` and infinity, respectively, for updates during optimization. This setup ensures the algorithm is ready to iteratively evolve solutions based on the fitness function and particle interactions.

In \textbf{\textit{Algorithm \ref{empso_opti}}}, we define the EMPSO Optimization function, which incorporates the fitness function, bounds, number of particles, momentum factor, and cognitive/social learning parameters. The optimization function iteratively updates particle velocities and positions based on their personal bests and the global best, ultimately returning the optimized perturbations $\mathbf{v}^*_{\text{gbest}}$, which is equal to \( \mathbf{A}^* \).

\begin{algorithm}[H]
\caption{EMPSO Optimization}
\label{empso_opti}
\begin{algorithmic}[1]
\STATE \textbf{Input:} $T_{\text{max}}$, $\beta$, $c_1$, $c_2$, $\mathbf{b}$
\STATE \textbf{Output:} $\mathbf{v}^*_{\text{gbest}}$ (global best velocity), $f^*_{\text{gbest}}$ (global best fitness)
\FOR{$t = 1$ to $T_{\text{max}}$}
    \FOR{each particle $i \in S$ (swarm)}
        \STATE Evaluate fitness: $f_i \leftarrow f(\mathbf{x}_i)$
        \STATE Update personal best: $f_{\text{pbest}_i} \leftarrow \min(f_{\text{pbest}_i}, f_i)$
        \IF{$f_i < f^*_{\text{gbest}}$}
            \STATE Update global best value: $f^*_{\text{gbest}} \leftarrow f_i$
            \STATE Update global best velocity: $\mathbf{v}^*_{\text{gbest}} \leftarrow \mathbf{v}_i$
        \ENDIF
    \ENDFOR
    \FOR{each particle $i \in S$}
        \STATE Generate random values: $r_1, r_2 \sim \mathcal{U}(0, 1)$ (Uniform distribution)
        \STATE Update velocity:
        \[
        \mathbf{v}_i \leftarrow \beta \mathbf{m}_i + (1 - \beta) \mathbf{v}_i + c_1 r_1 (\mathbf{v}_{\text{pbest}_i} - \mathbf{x}_i) + c_2 r_2 (\mathbf{v}^*_{\text{gbest}} - \mathbf{x}_i)
        \]
        \STATE Update position: $\mathbf{x}_i \leftarrow \mathbf{x}_i + \mathbf{v}_i$
        \STATE Update momentum: $\mathbf{m}_i \leftarrow \beta \mathbf{m}_i + (1 - \beta) \mathbf{v}_i$
        \STATE Clip position within bounds: $\mathbf{x}_i \leftarrow \text{clip}(\mathbf{x}_i, \mathbf{b}_{\text{min}}, \mathbf{b}_{\text{max}})$
    \ENDFOR
\ENDFOR
\STATE \textbf{return} $\mathbf{v}^*_{\text{gbest}}$, $f^*_{\text{gbest}}$
\end{algorithmic}
\end{algorithm}

The optimization process involves the following steps:

\subsubsection{Initialization}
The algorithm begins by initializing a population of particles. Each particle represents a potential perturbation $\mathbf{v}_i$ in the search space, where the velocity of a particle corresponds to a candidate perturbation. The algorithm requires several input parameters: 
\begin{itemize}
    \item $T_{\text{max}}$ — the maximum number of iterations.
    \item $\beta$ — the momentum factor, which influences the previous velocity in the current velocity calculation.
    \item $c_1$ and $c_2$ — cognitive and social learning parameters, respectively.
    \item $\mathbf{b}$ — bounds that restrict the particles' movement within a defined search space.
\end{itemize}

\subsubsection{Fitness Evaluation}
At the start of each iteration, the algorithm evaluates the fitness of each particle $i$ in the swarm. The fitness function $f(\mathbf{v}_i)$ computes how well the perturbation represented by $\mathbf{v}_i$ performs in terms of the learner's classification error. The fitness score for each particle is compared to its personal best, denoted as $f_{\text{pbest}_i}$. If the new fitness is better (lower), the particle's personal best is updated.

\subsubsection{Global Best Update}
In addition to personal best solutions, the global best solution across the entire swarm is tracked. If a particle achieves a fitness value that is better than the current global best $f_{\text{gbest}}$, both the global best fitness $f^*_{\text{gbest}}$ and the global best velocity $\mathbf{v}^*_{\text{gbest}}$ are updated to reflect the particle's current fitness and velocity.

\subsubsection{Velocity and Position Update}
For each particle $i$, the algorithm generates two random values $r_1$ and $r_2$ from a uniform distribution, which introduce randomness into the search process, ensuring the exploration of the search space. The velocity update is a weighted sum of three components:
\begin{itemize}
    \item The momentum term, controlled by the factor $\beta$, which influences the particle to continue its previous motion.
    \item The cognitive term, driven by the difference between the particle's personal best velocity and its current position.
    \item The social term, driven by the difference between the global best velocity and the particle's current position.
\end{itemize}
The new velocity $\mathbf{v}_i$ is computed as:
\[
\mathbf{v}_i \leftarrow \beta \mathbf{m}_i + (1 - \beta) \mathbf{v}_i + c_1 r_1 (\mathbf{v}_{\text{pbest}_i} - \mathbf{x}_i) + c_2 r_2 (\mathbf{v}^*_{\text{gbest}} - \mathbf{x}_i)
\]
Once the velocity is updated, the particle's position $\mathbf{x}_i$ is updated by adding the velocity to the current position, and the momentum $\mathbf{m}_i$ is also updated.

\subsubsection{Position Clipping}
To ensure that particles do not exceed the search bounds, the position of each particle is clipped within predefined bounds $\mathbf{b_{\text{min}}}$ and $\mathbf{b_{\text{max}}}$, ensuring that the perturbations remain valid within the defined problem space.

\subsubsection{Termination and Output}
The algorithm repeats the process of evaluating fitness, updating velocities, and updating positions for a total of $T_{\text{max}}$ iterations. Upon completion, the global best velocity $\mathbf{v}^*_{\text{gbest}}$ is returned as the optimized perturbation(\( \mathbf{A}^* \)), which is the best solution found for maximizing the learner's classification error.

The EMPSO optimization process ensures that adversarial perturbations are refined over iterations, allowing the adversary to generate effective perturbations that significantly affect the model's performance. The iterative nature of the algorithm helps balance exploration of new perturbations and exploitation of the best-found solutions, ultimately converging toward an optimal perturbation for adversarial attacks. To validate the effectiveness of the proposed method, we now turn to the experimental setup and results in Sec. \ref{exp}.

\section{Experiments}
\label{exp}
In this section, we conduct experiments to validate the hypothesis\textit{(see Box 1)}. that the adversary improves its ability to produce effective perturbations, thereby reducing the learner's classification accuracy and revealing weaknesses in its decision boundaries.

During the training process, the adversary's objective is to discover adversarial manipulations \( \mathbf{A}^* \) that cause the learner model to misclassify positive class examples. The learner is then retrained using these adversarial examples to enhance its robustness. We evaluate the proposed ExAL algorithm on two datasets: MNIST Handwritten Digits (\citet{lecun1998mnist}) and the Blended Malware Dataset (\citet{pendharkar2021blended}). For each dataset, experiments are performed on binary classification tasks, where adversarial data is generated and analyzed.

\subsection{Experiment Setup}
For both experiments, the dataset is filtered to include pairs of class labels as shown in the \textit{Labels} column of Tables \ref{tab:mnist_f1} and \ref{tab:malware_performance}, respectively. For each pair of labels, the left label is designated as the positive class, while the right label represents the negative class. To accommodate computational resource constraints, only 1,000 samples are selected per class. The setup for these experiments is summarized in Box 3, while the overall experimental procedure is outlined in Box 4.

Adversarial data is generated by training the adversary using the ExAL algorithm to produce optimal perturbations, denoted as \( \mathbf{A}^* \). These perturbations are applied to the original data to create perturbed examples. The perturbations are designed to cause the model to misclassify positive examples as negative and vice versa.

\begin{figure}
    \centering
    \includegraphics[width=\textwidth]{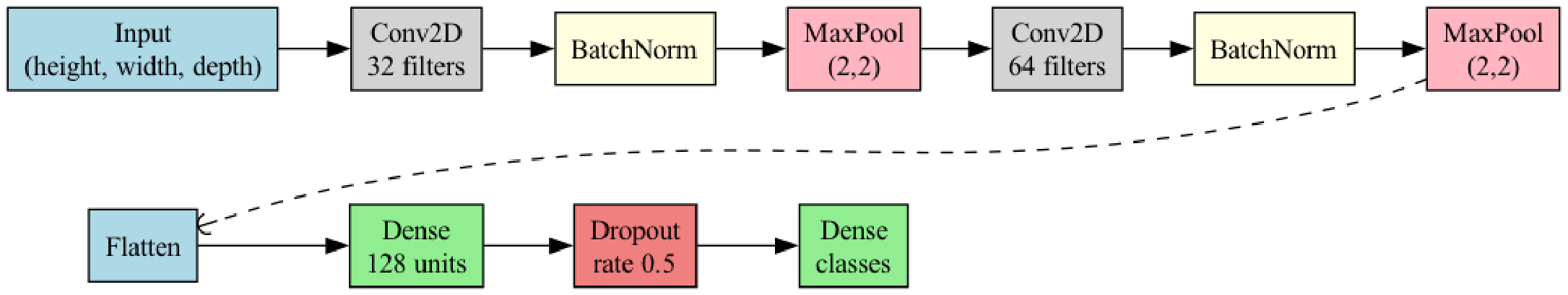}
    \caption{CNN Model Architecture}
    \label{cnn_model_mnist}
\end{figure}

\begin{tcolorbox}[colframe=blue!50!black, colback=blue!10!white, coltitle=white, title=Box 3: Experiment Setup, fonttitle=\bfseries]
For all the experiments we have the following:
\begin{enumerate}
    \item One independent adversary attacks the learner.
    \item $CNN_{original}$, $CNN_{manipulated}$ and $CNN_{secure}$ models are defined according to Definitions \ref{def:cnn_original}, \ref{def:cnn_manipulated} and \ref{def:cnn_secure} respectively.
    \item The model architecture is as depicted in Fig. \ref{cnn_model_mnist}.
\end{enumerate}
\end{tcolorbox}

\begin{tcolorbox}[colframe=green!50!black, colback=green!10!white, coltitle=white, title=Box 4: Experiment Steps, fonttitle=\bfseries]
\label{box:experiment_steps}
Each experiment proceeds through the following sequence of steps:
\begin{enumerate}
    \item Load the dataset and preprocess the data.
    \item Run Two-Player-Game using ExAL to generate adversarial data.
    \item Train and evaluate the F1-scores for \( \text{CNN}_{\text{normal}} \), \( \text{CNN}_{\text{manipulated}} \), and \( \text{CNN}_{\text{secure}} \).

\end{enumerate}
\end{tcolorbox}

\subsection{Experiment 1: MNIST Dataset}
\label{exp1}
In this experiment, the MNIST Handwritten Digits dataset is used for binary classification tasks. For example, in the label pair (2, 8), the digit \textit{2} is treated as the positive class, while the \textit{8} is treated as the negative class. Adversarial data is generated by applying the perturbations \( \mathbf{A}^* \) to the original images, creating perturbed images. The leftmost image of Fig. \ref{fig:mnist} shows the original images, the center image displays the generated perturbations, and the rightmost image presents the perturbed images. The results of this experiment, in terms of F-1 scores for the three models ($CNN_{\text{normal}}$, $CNN_{\text{manipulated}}$, and $CNN_{\text{secure}}$), are reported in Table \ref{tab:mnist_f1}.

\subsection{Experiment 2: Malware Dataset}
\label{exp2}
In this experiment, the Blended Malware Dataset is used for binary classification tasks. For instance, in the label pair (Fasong, Dinwod), the malware \textit{Fasong} is considered the positive class, while \textit{Dinwod} is treated as the negative class. Similar to the MNIST experiment, adversarial perturbations \( \mathbf{A}^* \) are generated and applied to the original samples, creating perturbed examples. The results of this experiment, measured in terms of F-1 scores, are shown in Table \ref{tab:malware_performance}.

\section{Results and Discussion}
\label{results_discussion}
In this section, we analyze the experimental outcomes to validate the hypothesis\textit{(see Box 1)} that the adversary improves its ability to produce effective perturbations, systematically reducing the learner's classification accuracy and revealing weaknesses in its decision boundaries. 

In both experiments, adversarial examples generated using the ExAL algorithm effectively reduced the classification accuracy of the learner model, demonstrating the adversary's capability to exploit weaknesses in the decision boundaries. Conversely, the performance of secure model, trained on adversarially perturbed examples, surpassed that of manipulated model, illustrating the effectiveness of adversarial training in improving model robustness. 

\begin{figure}[H]
    \centering
    \begin{subfigure}[b]{0.7\textwidth}
        \centering
        \includegraphics[width=\textwidth]{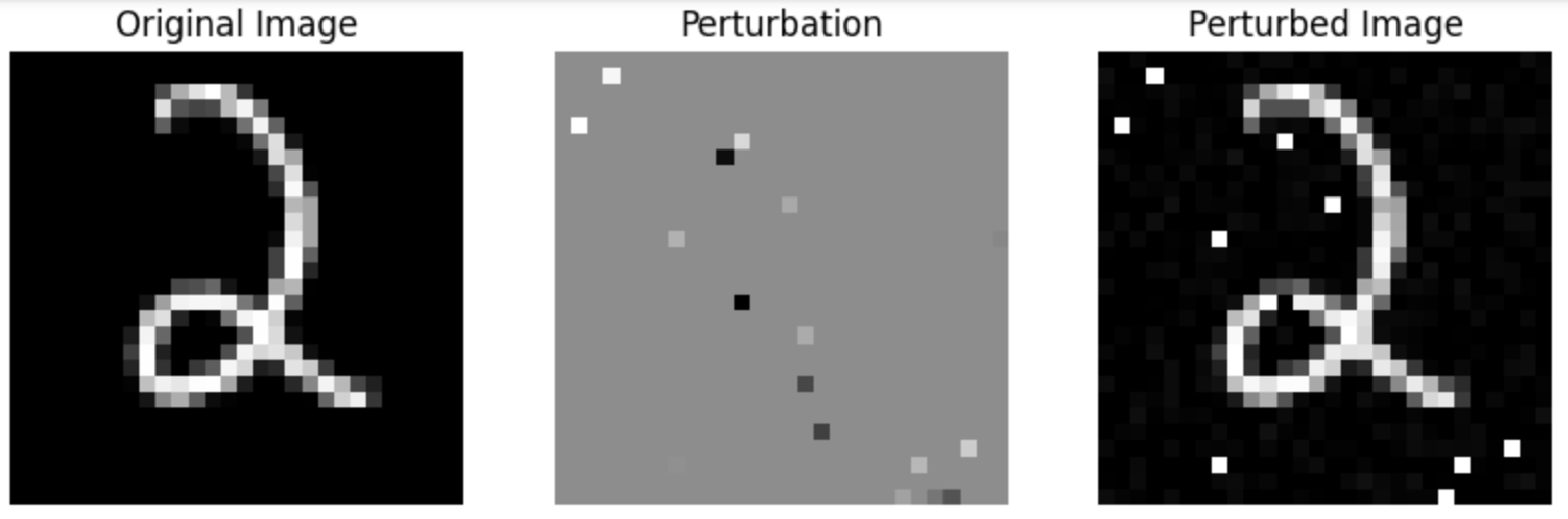}
        \caption{Label 2}
        \label{fig:image1}
    \end{subfigure}
    
    \begin{subfigure}[b]{0.7\textwidth}
        \centering
        \includegraphics[width=\textwidth]{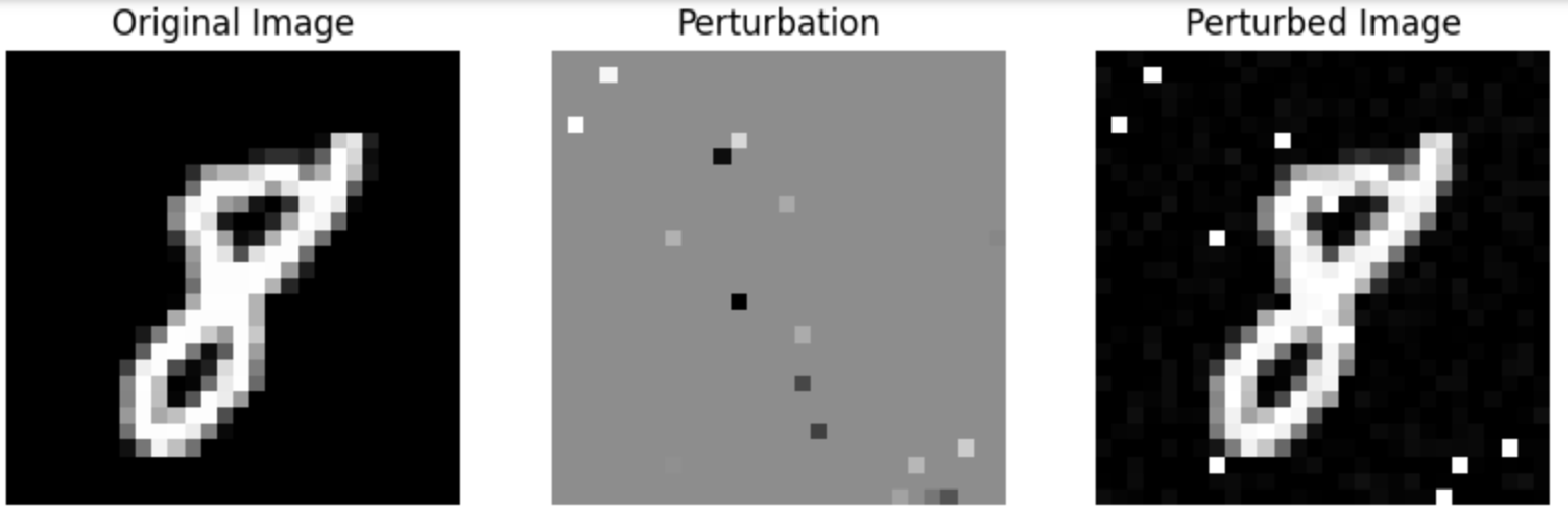}
        \caption{Label 8}
        \label{fig:image2}
    \end{subfigure}
    \caption{Original image(left), Adversarial Perturbation with scale factor 1 (center), Perturbed Image(Right)}
    \label{fig:mnist}
\end{figure}

For Experiment 1 in Sec. \ref{exp1} on the MNIST dataset, the perturbations generated by ExAL look like minor pixel-level changes as seen in Fig. \ref{fig:mnist}. Despite their subtlety, these perturbations significantly deceived the $CNN_{\text{manipulated}}$ model, leading to mis-classification and a decline in its performance, as reflected in its lower F1-scores in Table \ref{tab:mnist_f1}. The robustness of $CNN_{\text{secure}}$ was evident in its ability to maintain higher F1-scores compared to $CNN_{\text{manipulated}}$ across all label pairs.

\begin{table}[H]
  \caption{Original, Manipulated and Secure model performance on MNIST}
  \label{tab:mnist_f1}
  \centering
  \begin{tabular}{lllll}
    \toprule
    \multicolumn{3}{c}{\textbf{F1-scores}} &
    \textbf{Hypothesis} &
    \textbf{Labels} \\
    \cmidrule(r){1-3}
    $CNN_{org}$ & $CNN_{manip}$ & $CNN_{sec}$ & & \\
    \midrule
    0.8991 & 0.8798 & \textbf{0.9759} & Satisfied & (2,8) \\
    0.8801 & 0.3878 & \textbf{0.8811} & Satisfied & (4,9) \\
    \textbf{0.9507} & 0.4122 & 0.8854 & Satisfied & (1,4) \\
    0.8761 & 0.8628 & \textbf{0.8838} & Satisfied & (7,9) \\
    0.9548 & 0.9278 & \textbf{0.9699} & Satisfied & (6,8) \\
    \textbf{0.9721} & 0.8437 & 0.9469 & Satisfied & (2,6) \\
    \bottomrule
  \end{tabular}
\end{table}

A similar pattern was observed in Experiment 2 in Sec. \ref{exp2} on the Malware dataset, where the adversarial perturbations were generated to misclassify malware samples. The perturbed images caused a marked decline in the performance of $CNN_{\text{manipulated}}$. This is evident from its lower F1-scores in Table \ref{tab:malware_performance}. Conversely, $CNN_{\text{secure}}$ demonstrated a significant improvement, achieving higher F1-scores and showcasing its resilience against adversarial attacks.

\begin{table}
  \caption{Original, Manipulated and Secure model performance on Malware Dataset}
  \label{tab:malware_performance}
  \centering
  \begin{tabular}{llllll}
    \toprule
    \multicolumn{3}{c}{\textbf{F1-scores}} &
    \textbf{Hypothesis} &
    \textbf{Labels} & \textbf{Scale} \\
    \cmidrule(r){1-3}
    $CNN_{org}$ & $CNN_{manip}$ & $CNN_{sec}$ & & & \\
    \midrule
    1.0000 & 0.9857 & \textbf{1.0000} & Satisfied & (Fasong, Dinwod)        & 1   \\
    1.0000 & 0.2644 & \textbf{1.0000} & Satisfied & (VBA, VBKrypt)           & 5   \\
    1.0000 & 0.7097 & \textbf{1.0000} & Satisfied & (Fakerean, Autorun)      & 5   \\
    1.0000 & 0.9923 & \textbf{1.0000} & Satisfied & (InstallCore, BrowseFox) & 5   \\
    1.0000 & 0.4039 & \textbf{1.0000} & Satisfied & (Adposhel, Amonetize)    & 5   \\
    0.9537 & 0.9458 & \textbf{0.9653} & Satisfied & (Injector, Androm)       & 0.5 \\
    \bottomrule
  \end{tabular}
\end{table}

\begin{figure}
    \centering
    \begin{subfigure}[b]{0.7\textwidth}
        \centering
        \includegraphics[width=\textwidth]{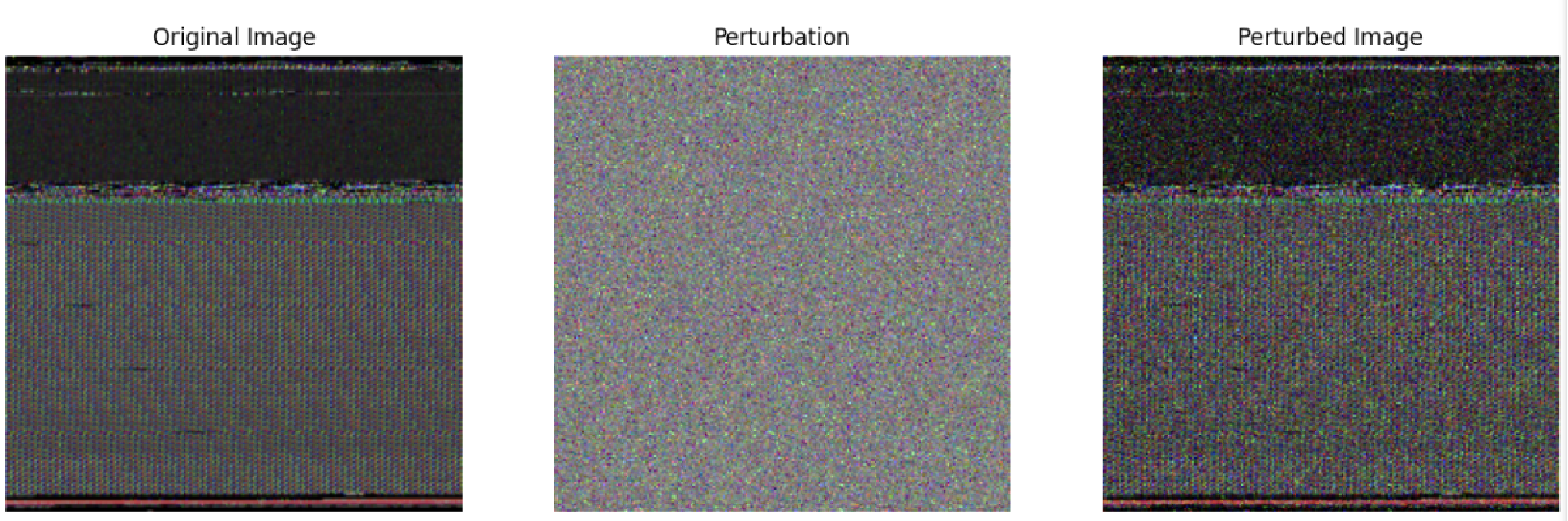}
        \caption{Label VBA}
        \label{fig:image5}
    \end{subfigure}
    
    \begin{subfigure}[b]{0.7\textwidth}
        \centering
        \includegraphics[width=\textwidth]{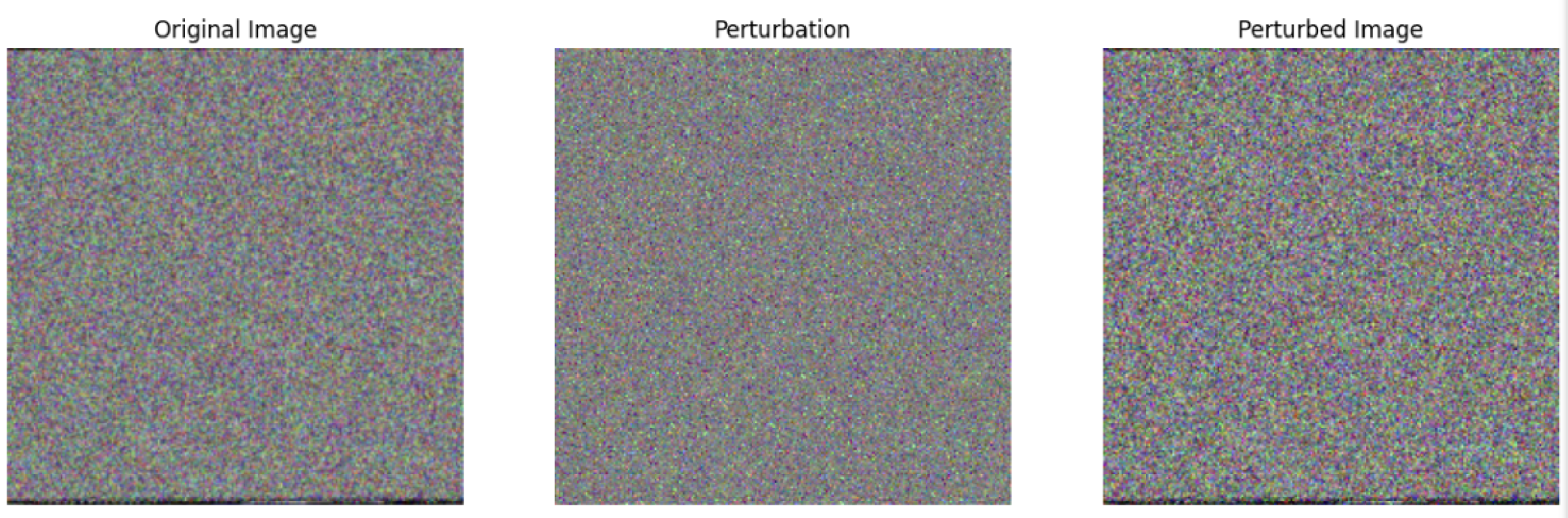}
        \caption{Label VBAKrypt}
        \label{fig:image6}
    \end{subfigure}
    \caption{Original image(left), Adversarial Perturbation with scale factor 5 (center), Perturbed Image(Right)}
    \label{fig:malware2}
\end{figure}

\begin{figure}
    \centering
    \begin{subfigure}[b]{0.7\textwidth}
        \centering
        \includegraphics[width=\textwidth]{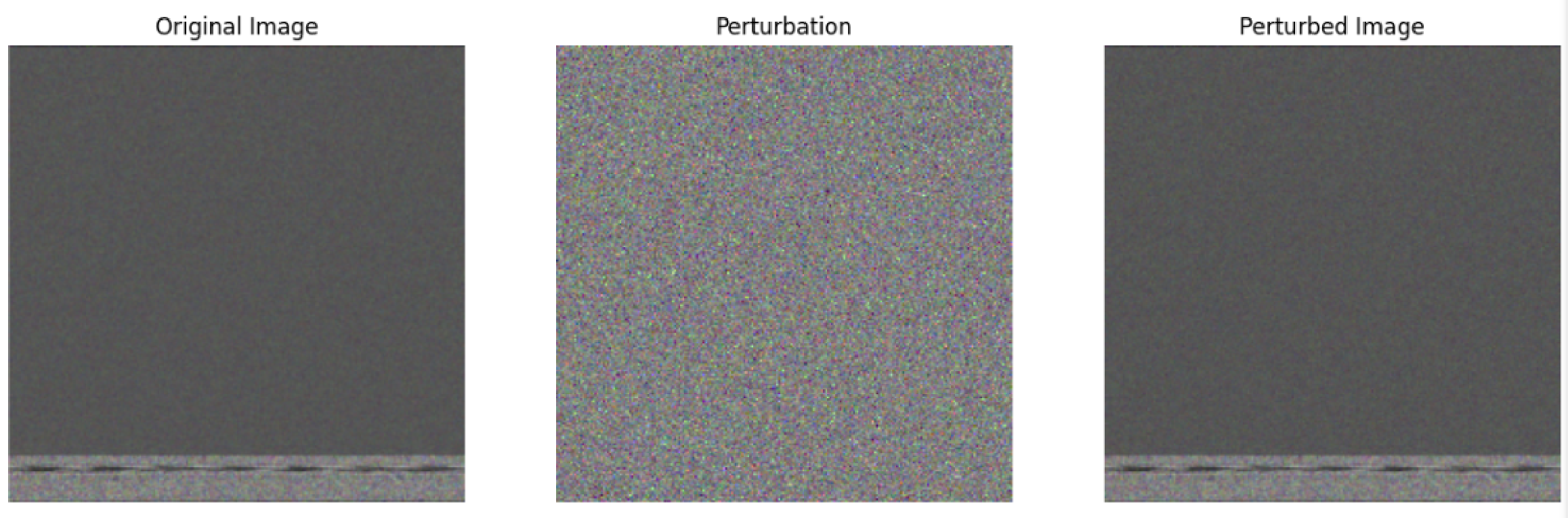}
        \caption{Label Injector}
        \label{fig:image7}
    \end{subfigure}
    
    \begin{subfigure}[b]{0.7\textwidth}
        \centering
        \includegraphics[width=\textwidth]{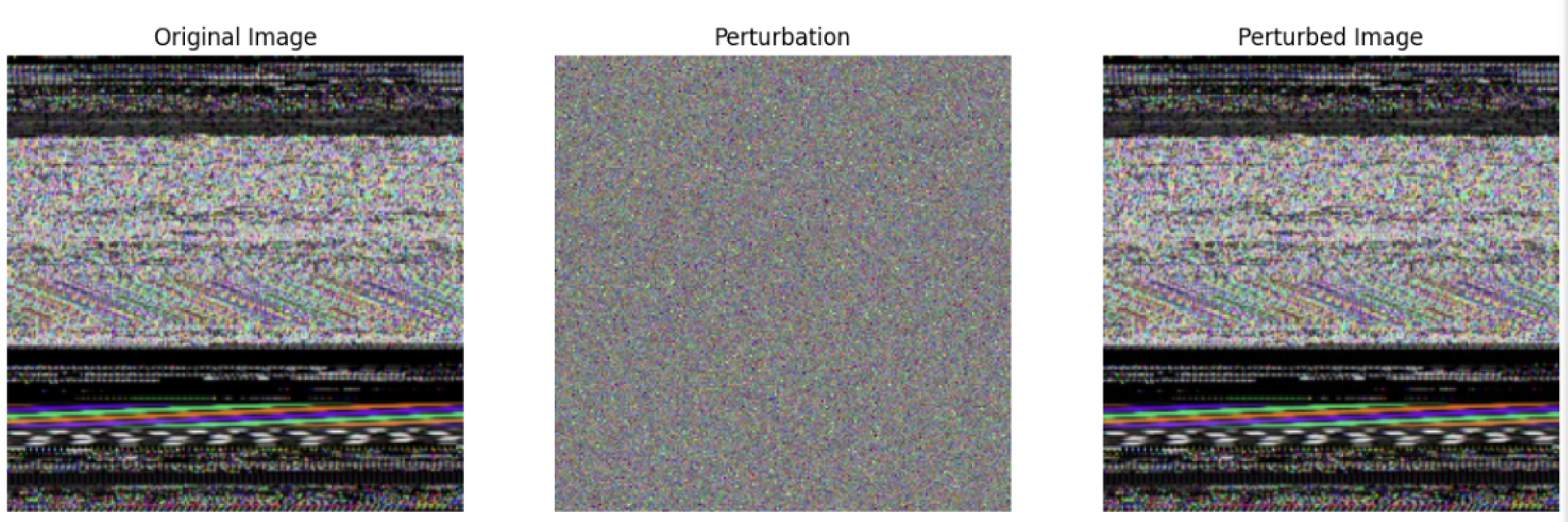}
        \caption{Label Androm}
        \label{fig:image8}
    \end{subfigure}
    \caption{Original image(left), Adversarial Perturbation with scale factor 0.5 (center), Perturbed Image(Right)}
    \label{fig:malware3}
\end{figure}

To evaluate the impact of perturbation strength, we experimented with different scale factors. With a lower scale factor of 0.5, the adversarial perturbations were milder (see Fig. \ref{fig:malware3}). In this scenario, $CNN_{\text{manipulated}}$ showed marginally improved performance but still lagged behind $CNN_{\text{secure}}$. When the scale factor was increased to 5, the generated perturbations (Fig. \ref{fig:malware2}) remained visually indistinguishable but caused a drastic decline in $CNN_{\text{manipulated}}$'s performance. Meanwhile, $CNN_{\text{secure}}$ achieved near-perfect F1-scores, as observed in Table \ref{tab:malware_performance}, further demonstrating its robustness to stronger adversarial attacks.

These findings validate the efficacy of the ExAL algorithm in generating optimal perturbations that are both visually indistinguishable and capable of significantly impairing model performance. Moreover, they highlight the robustness of $CNN_{\text{secure}}$, trained on these adversarial examples, in securing the model against further perturbations. The hypothesis\textit{(see Box 1)} is validated in all the cases as shown in Tables \ref{tab:mnist_f1} and \ref{tab:malware_performance}, confirming that adversarial training with ExAL effectively enhances model defenses against adversarial attacks.

\section{Conclusion and Future Work}
\label{concl}

In this paper, we introduce ExAL, an adversarial learning framework grounded in the EMPSO optimization paradigm, aimed at enhancing the robustness of deep learning models under adversarial scenarios. ExAL emphasizes exploration within the search landscape to effectively identify adversarial perturbations that significantly compromise model performance. By leveraging these perturbations in an adversarial training loop, ExAL equips the learner model with enhanced resilience to subsequent adversarial attacks. We validate the effectiveness of ExAL through experiments on CNN-based architectures, demonstrating its ability to generate impactful perturbations while securing the model against adversarial vulnerabilities. In addition to outperforming baseline models, ExAL illustrates its adaptability to varying perturbation magnitudes, highlighting its robustness across multiple experimental setups. 

For future work, we aim to expand the application of ExAL to more advanced architectures, including Kolmogorov-Arnold Networks (KAN). Integrating ExAL with KAN holds promise for uncovering novel insights, particularly in domains requiring interpretable and efficient models. Additionally, we intend to explore a quantum variant of ExAL, leveraging quantum computing principles to further enhance its optimization capabilities and adapt it to quantum machine learning architectures.

\begin{credits}
\subsubsection{\ackname} To Anish Kumar Kallepalli, for providing the base code for the adversarial learning game. 

\subsubsection{\discintname}
\textbf{The authors have no competing interests to declare that are
relevant to the content of this article.}
\end{credits}
%
%
%
%

\section*{CRediT Authorship Contribution Statement}

\noindent\fbox{
\begin{minipage}{\textwidth}
\centering

\renewcommand{\arraystretch}{1.5} 
\begin{tabularx}{\textwidth}{>{\bfseries}X|X}
\hline
\rowcolor{gray!20} \textbf{Contribution} & \textbf{Author(s)} \\ \hline
\rowcolor{gray!10} Conceptualization & A Vinil, Aneesh Sreevallabh Chivukula \\ 
\rowcolor{white} Methodology & A Vinil \\ 
\rowcolor{gray!10} Software & A Vinil \\ 
\rowcolor{white} Writing - Original Draft & A Vinil \\ 
\rowcolor{gray!10} Writing - Review \& Editing & A Vinil, Aneesh Sreevallabh Chivukula \\ 
\rowcolor{white} Supervision & Aneesh Sreevallabh Chivukula \\ 
\rowcolor{gray!10} Algorithm Research & Aneesh Sreevallabh Chivukula \\  
\rowcolor{white} Application Research in Malware Analysis & Pranav Chintatreddy \\ 
\rowcolor{gray!10} Data Curation & A Vinil, Pranav Chintatreddy \\ 
\hline
\end{tabularx}
\end{minipage}
}

\bibliographystyle{apalike}
\bibliography{bibfile}
\end{document}